\documentclass[conference]{IEEEtran}
\IEEEoverridecommandlockouts
\usepackage{cite}
\usepackage{amssymb,amsfonts}
\usepackage{algorithmic}
\usepackage{amsmath}
\usepackage{graphicx}
\usepackage{textcomp}
\usepackage{svg}
\usepackage{xcolor}
\usepackage{listings}
\usepackage{url}
\def\BibTeX{{\rm B\kern-.05em{\sc i\kern-.025em b}\kern-.08em
		T\kern-.1667em\lower.7ex\hbox{E}\kern-.125emX}}
\definecolor{light-gray}{gray}{0.95}
\newcommand{\code}[1]{\colorbox{light-gray}{\texttt{#1}}}
\PassOptionsToPackage{hyphens}{url}\usepackage[hidelinks]{hyperref}

\begin{document}
	
	\title{Interactive Learning in Computer Science Education Supported by a Discord Chatbot}
	
	\author{\IEEEauthorblockN{
        \begin{minipage}{0.4\textwidth}
            \centering
            Santiago Berrezueta-Guzman\\
            \textit{Technical University of Munich}\\
            Heilbronn, Germany \\
            s.berrezueta@tum.de
        \end{minipage}
        \hspace{0.1\textwidth}
        \begin{minipage}{0.4\textwidth}
            \centering
            Ivan Parmacli\\
            \textit{Technical University of Munich}\\
            Heilbronn, Germany \\
            ivan.parmacli@tum.de
        \end{minipage}
    }
    \vspace{1\baselineskip} 
    \IEEEauthorblockN{
        \begin{minipage}{0.4\textwidth}
            \centering
            Stephan Krusche\\
            \textit{Technical University of Munich}\\
            Munich, Germany \\
            krusche@tum.de
        \end{minipage}
        \hspace{0.1\textwidth}
        \begin{minipage}{0.4\textwidth}
            \centering
            Stefan Wagner\\
            \textit{Technical University of Munich}\\
            Heilbronn, Germany \\
            stefan.wagner@tum.de
        \end{minipage}
    }
	}
	
	\maketitle
	
	\begin{abstract}
		
Enhancing interaction and feedback collection in a first-semester computer science course poses a significant challenge due to students' diverse needs and engagement levels. To address this issue, we created and integrated a command-based chatbot on the course communication server on Discord. The DiscordBot enables students to provide feedback on course activities through short surveys, such as exercises, quizzes, and lectures, facilitating stress-free communication with instructors. It also supports attendance tracking and introduces lectures before they start.

The research demonstrates the effectiveness of the DiscordBot as a communication tool. The ongoing feedback allowed course instructors to dynamically adjust and improve the difficulty level of upcoming activities and promote discussion in subsequent tutor sessions. The data collected reveal that students can accurately perceive the activities' difficulty and expected results, providing insights not possible through traditional end-of-semester surveys. Students reported that interaction with the DiscordBot was easy and expressed a desire to continue using it in future semesters. This responsive approach ensures the course meets the evolving needs of students, thereby enhancing their overall learning experience.
		
	\end{abstract}
	
	
	\section{Introduction}
	Continuous feedback from students in introductory computer science courses (CS1) is vital for the development and effectiveness of education in this area. It allows educators to adjust teaching methods and course activities to match the rapidly changing demands of the field, thus enhancing student learning outcomes \cite{merelo2023chatbots}. This feedback is crucial for promptly identifying first-year students' challenges, reducing their frustration and likelihood of dropping out \cite{ramu2023generation}. Additionally, students' feedback promotes a learner-centered approach, increasing engagement and motivation, and contributes to a more inclusive and adaptable educational environment \cite{chempavathy2022ai}.
	
	On the other hand, Chatbots\footnote{A chatbot is a bot designed to simulate human-like conversations with users via the internet, responding to queries and interacting in real-time \cite{adamopoulou2020overview}.} are transforming educational procedures by offering interactive and personalized learning experiences. These bots are tools powered or operated by artificial intelligence (AI-driven assistants) and provide on-demand support, enabling students to get immediate answers and engage more effectively, regardless of time or location \cite{annuvs2023chatbots}. By integrating chatbots, lecture platforms become more dynamic and responsive, enhancing the traditional learning process. Furthermore, chatbots can facilitate a better understanding of course material by providing students with instant feedback and supplementary resources, thereby augmenting the conventional lecture experience \cite{riza2023use}, \cite{adiguzel2023revolutionizing}.
	
	In this article, we introduce a Discord\footnote{Discord is a voice, video, and text chat app used to talk and hang out with their communities and friends, \url{https://discord.com/}.} chatbot that we called DiscordBot and helps to collect continuous feedback from students on various aspects of an introductory programming course. This feedback includes the difficulty of quizzes, lectures, exercises, and evaluations to effectively improve these activities and teaching methods. Additionally, the chatbot assists in disseminating lecture materials and tracking attendance during tutorials.
	
	This paper is composed of an upcoming section \ref{RW} that reviews existing research and a section \ref{M} that outlines the methods used to develop and evaluate this DiscordBot. Section \ref{R} details the results obtained in the integration, which are explored further in section \ref{D}. Finally, the conclusions and future directions are discussed in section \ref{C}. 
	
	\section{Related work}\label{RW}
	
	In education, chatbots have emerged as innovative tools for learning support, offering personalized assistance and interactive communication to foster a more engaging and practical educational experience. They can be used in various educational settings, from traditional classrooms to online courses. They can assist with a wide range of tasks, such as answering student queries, providing feedback on assignments, and facilitating group discussions. 

     Previous research has shown that Discord significantly enhances engagement and community building in online higher education, critical during the COVID-19 pandemic when student engagement was crucial \cite{kruglyk2020discord, lauricella2023examining, ayob2022promoting, vladoiu2020learning}.
	
	Jung and Woo developed a chatbot to improve computer-supported collaborative learning by providing educational, social, managerial, and technical support. The chatbot was evaluated for usability and significantly boosted collaborative activities and user engagement. These results underscore the chatbot's role in fostering an interactive and supportive learning environment, illustrating its potential benefits in modern educational settings \cite{jung2021developing}.
	
	Gowda et al. integrated a rule-based chatbot into Discord, exploring how such technologies, driven by artificial intelligence (AI) and natural language processing (NLP), are changing education. The study focused on developing a rule-based chatbot to address educational challenges. The chatbot's standout feature is its Q/A functionality, which pulls extensive information from preset data, enhancing teaching and learning experiences by facilitating interactive conversations. This research highlights NLP's critical role in advancing chatbot capabilities \cite{cp2021development}.
	
	Unlike previous models that mainly addressed common queries, DiscordBot\footnote{DiscordBot source code: \url{https://github.com/Johnypier/DiscordBot}} continuously collects detailed feedback from students about course activities. It asks students about the difficulty level of exercises and evaluations. Additionally, it promotes lecture attendance by presenting upcoming topics and facilitates the attendance tracking of tutorial sessions, making these processes more interactive. 
	
	\section{Methodology}\label{M}
	
	\subsection{Course selection}

    We chose an introductory programming course for the first semester of a computer science bachelor's program based on interactive learning \cite{krusche2017interactive, krusche2020interactive}. In this course, 137 students were registered, and the primary communication tool between fellow students and instructors is Discord. The course integrates onsite lectures, onsite tutorial sessions, quizzes at the beginning of each lecture \cite{krusche2023introduction}, weekly exercises on Artemis \cite{krusche2018artemis}, and computer-based exams \cite{linhuber2023constructive}. The course undergoes an end-semester evaluation that assesses the curriculum. However, it is not focused on collecting students' perceptions of individual course activities. Therefore, we considered it necessary to obtain continuous feedback from students to understand their perception of the difficulty of course activities such as exercises, quizzes, and evaluations and adapt them according to this feedback and the student's performance. 
    
    The proposed DiscordBot asks students for detailed feedback after each activity to address this gap, enabling instructors to make timely adjustments to these course activities and enhancing the learning experience. 
    Its functionality not only fosters continuous and specific feedback but also aids in tracking and encouraging active student engagement throughout the course. With this study, we plan to address two important research questions: 
	
	\begin{itemize}
		\item \textbf{RQ1}: Is the difficulty level perceived by the students about the activities (lectures and evaluations) related to the performance of the students? 
		\item \textbf{RQ2}: Is the difficulty level perceived by the students about the lecture content related to the difficulty level of the corresponding quiz? 
	\end{itemize}
	
	\subsection{DiscordBot - Architecture}
	
\begin{figure}[ht]
		\centering{\includegraphics[width=0.91\linewidth]{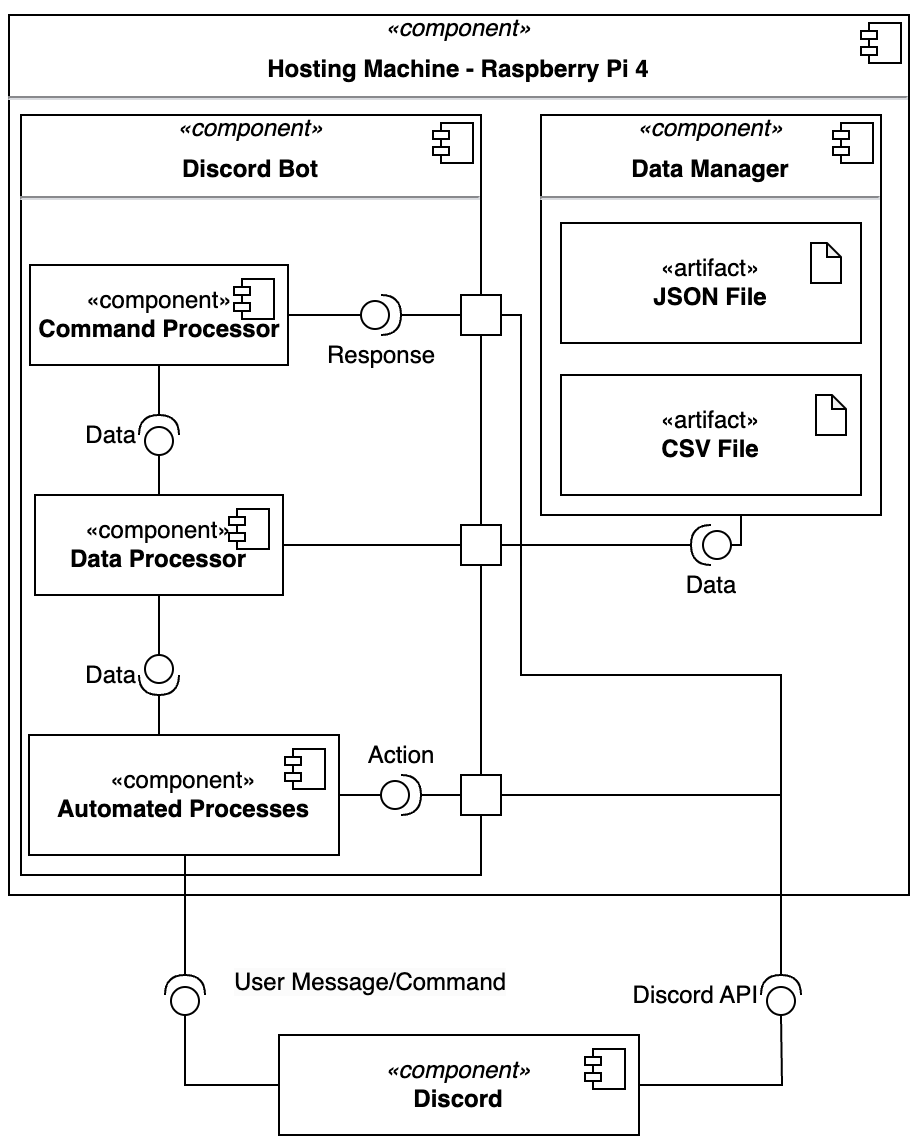}}
		\caption{Component diagram of the DiscordBot.}
		\label{ComponentDiagram}
	\end{figure}

	The DiscordBot is hosted on a Raspberry Pi model 4 in the university facility, running on the Debian operative system, and the course instructors maintain it.
	Figure \ref{ComponentDiagram} presents the main components of the DiscordBot. Its architecture is composed of three significant components: \textit{Discord},  \textit{Data Manager}, and  \textit{DiscordBot}.

	\textit{Discord} forwards the user's messages and commands to the \textit{DiscordBot} for processing and receives a response from it, which is reflected in the appropriate guild\footnote{In Discord, this represents an isolated collection of users and channels, and is often referred to as "server" in the user interface (UI).}.
	
	The \textit{Data Manager} allows course instructors to upload data to be processed, categorized, and saved in comma-separated value (CSV) or JavaScript Object Notation (JSON) files.
	
	The \textit{DiscordBot} uses the data by the Command Processor and Automated Processes components through the Data Processor component, which puts the data into a usable form to generate a response or perform a specific action.
	The Command Processor component generates a response based on a command or user message, and the Automated Processes component automatically performs specific actions at various times, which may be triggered by particular events in the guild or predetermined by data from instructors or developer instructions.
	
\subsection{Materials} 
	
	We used the \textit{Discord API}\footnote{Discord API repository,  \url{https://github.com/discord/discord-api-docs}} that is based on two core layers, a REST API that works around HTTP requests for general operations, and persistent secure Web-Socket\footnote{A Web-Socket is a computer communication protocol that enables full-duplex interaction between a web browser and a web server \cite{fette2011websocket}.} based connection for sending and subscribing to real-time events. 
	To authenticate a bot on Discord, the API provides a service or access to a platform through the OAuth2 API \footnote{With OAuth 2, an application can obtain a user's consent to call an API on their behalf, without needing the user's credentials for the API site \cite{wilson2022oauth}.}.
	
	The Rate Limits feature is integral to all Discord APIs and is designed to safeguard the platform from spam, abuse, and potential service overloads. These limits are meticulously enforced for each bot and user, with a dual-layered approach encompassing individual route-specific constraints and overarching global restrictions.
	Per-route rate limits are meticulously established for each route, encompassing numerous distinct endpoints. Additionally, these limits may pertain to various HTTP methods, including \code{GET}, \code{POST}, \code{PUT}, and \code{DELETE}. In certain instances, rate limits are collectively imposed on endpoints with analogous functionalities, ensuring a harmonized operational flow.
	While these mechanisms are pivotal in maintaining the integrity and reliability of the API, they can inadvertently impede the API's full utility. Although essential for preventing misuse, the imposition of rate limits requires a careful balance to allow legitimate and efficient use of the API's capabilities.
	
	Therefore, we used the Pycord API\footnote{Pycord documentation, \url{https://docs.pycord.dev/en/stable/}} to wrap the Discord API.
	This is a modern Pythonic API using the \textit{async/await} syntax. It implements the exact rate limit handling that prevents 429 HTTP status code problems and is easy to use with object-oriented programming designs.
	Pycord's commands extension simplifies bot creation to a few lines of code and makes registering new commands and events easy in Discord guilds. 
	
	After defining and implementing the command and event functions, the next critical step involves acquiring the Authentication Token to initiate the bot. This token is a unique identifier essential for the bot's operation.
	To generate this token, the 'Bot' section within the Discord application on the Discord Developer Portal presents the 'Reset' button. This button will create a new token that will be used to launch the bot. 
	
	\subsection{Student-DiscordBot Interactions}
	
\begin{figure}[ht]
		\centering{\includegraphics[width=\linewidth]{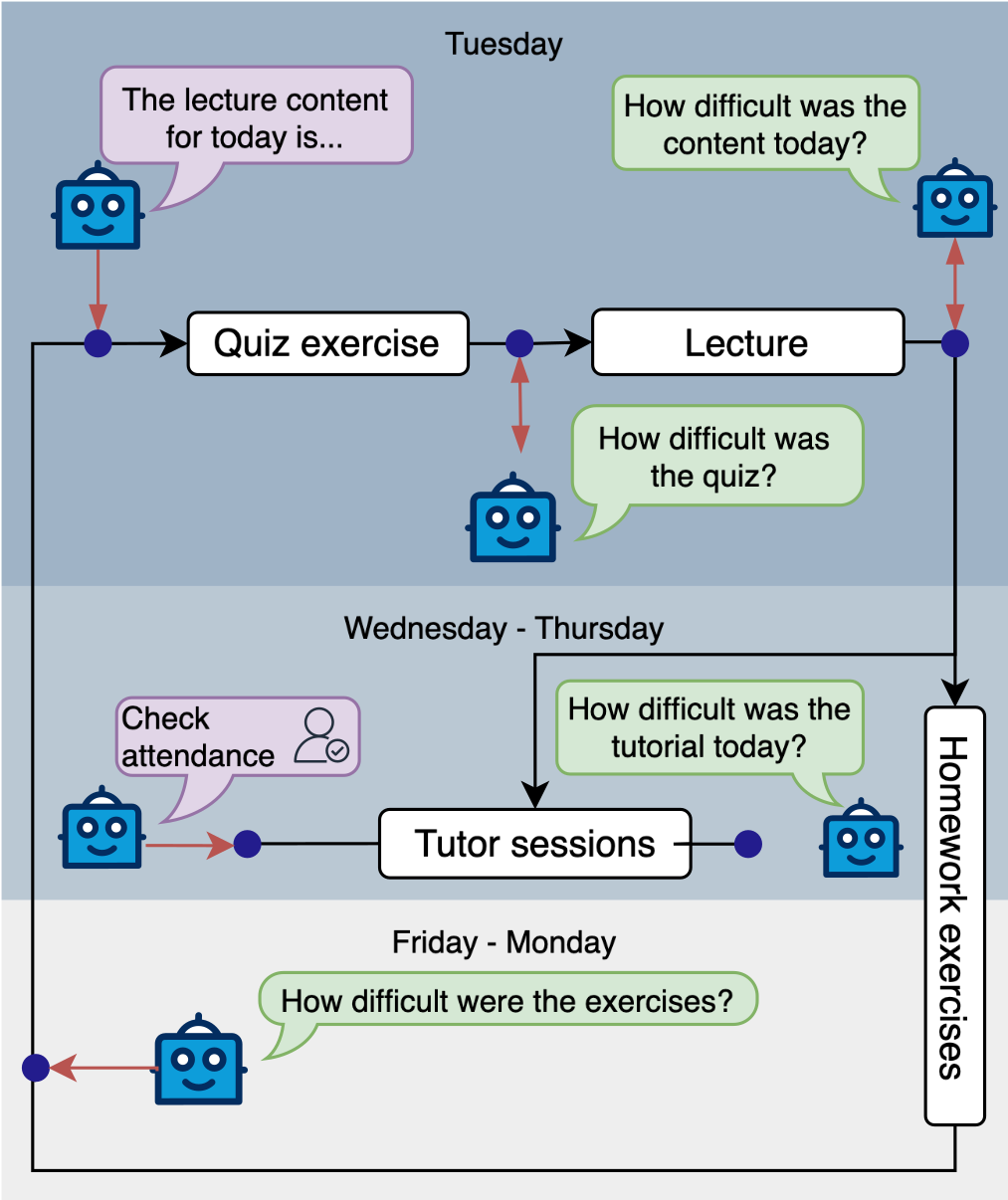}}
		\caption{Interaction of DiscordBot with students regarding course activities throughout the week. On Tuesday, DiscordBot invites students to attend the lecture, providing details on the corresponding content. It also inquires about the difficulty level of the quiz following the lecture and the lecture itself once it concludes. On Wednesday and Thursday, DiscordBot assists instructors in tracking attendance for tutorial sessions and gauging their difficulty level. After the weekly assignment deadline, it also collects student feedback.}
		\label{Flow}
	\end{figure}

	The introductory programming course features several activities throughout the week. Classes are held on Tuesdays and commence with a quiz exercise. Tutorial sessions take place from Wednesday to Thursday. Simultaneously, exercises are assigned for completion at home from Wednesday to Monday. Figure \ref{Flow} illustrates the process model of the course during the week and the precise moment when interaction with the DiscordBot occurs, facilitating information collection, notifications, or attendance checks.
	
	This interaction is facilitated through commands or by triggering events with particular keywords, which only the instructors can access. Discord has streamlined its app commands into a slash command format, where instructors need to type a slash (\code{/}) to access a list of supported commands within Discord.
	
\begin{figure}[ht]
		\centering{\includegraphics[width=0.8\columnwidth]{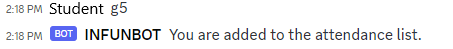}}
		\caption{Attendance Check interaction. The DiscordBot confirms that the student is in the attendance record after receiving the keyword "g5". Due to privacy restrictions, this screenshot does not contain the students' real names.}
		\label{attendance-response}
	\end{figure}

\begin{figure}[ht!]                                       
		\centering{\includegraphics[width=0.5\columnwidth]{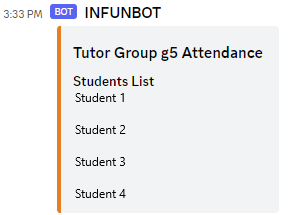}}
		\caption{Attendance Check bot-instructor interaction. The instructor receives a message with the list of the students in the session. Due to privacy restrictions, this screenshot does not contain the students' real names.}
		\label{attendance-instructor-response}
	\end{figure}

	The \textbf{Check Attendance interaction} enables instructors to quickly know who is present in the session; typically, this process takes less than one minute to complete. This interaction follows the steps outlined below:
	
	\begin{enumerate}
		\item The instructor can start and stop the check attendance process. 
		\item The instructor provides the students with a unique word (on-site) that will serve as a keyword. 
		\item Each student must answer the DiscordBot with the provided keyword. 
		\item The DiscordBot processes this input and displays a response to the student as shown in Figure \ref{attendance-response}. The students will know they are included in the list. 
		\item Automatically, the instructor can see a list of students who participated in the session, as shown in Figure \ref{attendance-instructor-response}. The check attendance list is also saved in a CSV file for further processing. 
	\end{enumerate}
	
\begin{figure}[ht]
		\centering{\includegraphics[width=\columnwidth]{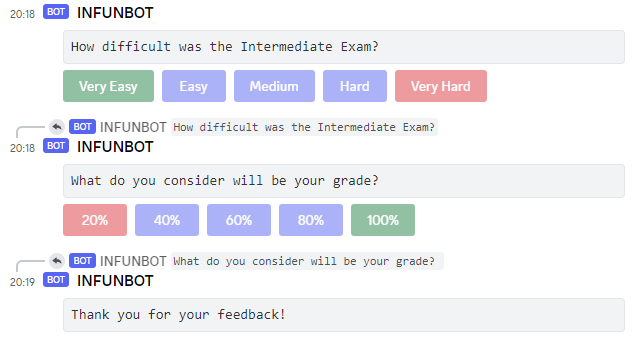}}
		\caption{Full interaction to collect feedback about an activity. The DiscordBot asked about the difficulty level of the intermediate exam after this ended and then asked about the estimated grade in a percentage range. The interaction ends with a "Thank you" message. }
		\label{survey-interaction}
	\end{figure}

\begin{figure}[ht]
		\centering{\includegraphics[width=0.8\columnwidth]{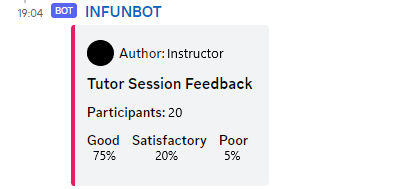}}
		\caption{Visualization of the collected tutor session feedback. The instructor has a global overview of the students' perception of the tutorial sessions. This allows us to address some modifications for future sessions.}
		\label{tutor-session}
	\end{figure}

	The \textbf{Feedback Collection interaction} with the students incorporates additional user interface (UI) elements, like buttons and selection menus. Depending on the activity (exercise, exam, evaluation, lecture, tutor session, or quiz), this interaction follows the steps outlined below:
	
	\begin{enumerate}
		\item The DiscordBot invites the students with a direct message to participate in the survey after the lecture, a quiz, an exercise, or an examination. 
		\item Students click the 'Accept' button to start the survey. 
		\item The bot sends a new message with the specific questions and the buttons to provide the answers.
		\item Once the interaction with the previous message is complete (the student has clicked a button), the buttons attached to that message are disabled. Thus, the student will know that the answer to the question has been saved (see Figure \ref{survey-interaction}).
		\item The instructor receives the final survey results (see Figure \ref{tutor-session}). However, this information has also been saved in a CSV file for further analysis. 
	\end{enumerate}
		
	Additionally, Discord supports Embed\footnote{A special message, often sent by a bot, can have a colored border, embedded images, text fields, and other fancy properties.} message creation, which allows the combination of custom user interface (UI) elements to create beautiful-looking interactions.
	Such an Embed is shown in Figure \ref{tutor-session}, where the required information is positioned; optionally, custom buttons can be attached.
	
	\subsection{Data Processing}
	
	The Matplotlib library (version 3.9.0) is used for data processing, and each interaction with the DiscordBot can be stored and categorized as needed. The records of students' attendance and feedback are preserved to assess student engagement and tutorial session performance. This data is saved in designated folders based on the activity. The DiscordBot stores the information in a CSV file, and instructors analyze it using a provided Python script that visualizes the data in charts. This data is only accessible to the course lecturer and instructors through direct access to the host machine, so students cannot access the data through interaction with the DiscordBot.

 \subsection{Usability and user satisfaction}

 At the conclusion of the course, we administered a comprehensive paper-based survey to the 116 students registered for the final exam. This survey was designed to evaluate the usability of DiscordBot, the frequency and quality of student interactions with the bot, the perceived usability of both Discord and the chatbot, and the accuracy of students' expected grades in each activity compared to their actual grades. Additionally, the survey gauged the students' intention to use DiscordBot in future courses.
	
	\section{Results}\label{R}
	
	This research aimed to determine whether using a Discord chatbot could positively impact students' enthusiasm for the subject and the development of the course as a whole.
	Throughout the course, we conducted a series of surveys using different versions of DiscordBot. These surveys covered various aspects of the course, including lectures, quizzes, exercises, exams, and study guides. The insights gathered from these surveys significantly influenced the course's direction and development.
	One of the most important outcomes was understanding how well we assessed student proficiency when we created the exam, how challenging the tasks were, and what results students expected.
	
\begin{figure}[ht]
		\centering
		\includegraphics[width=0.96\columnwidth]{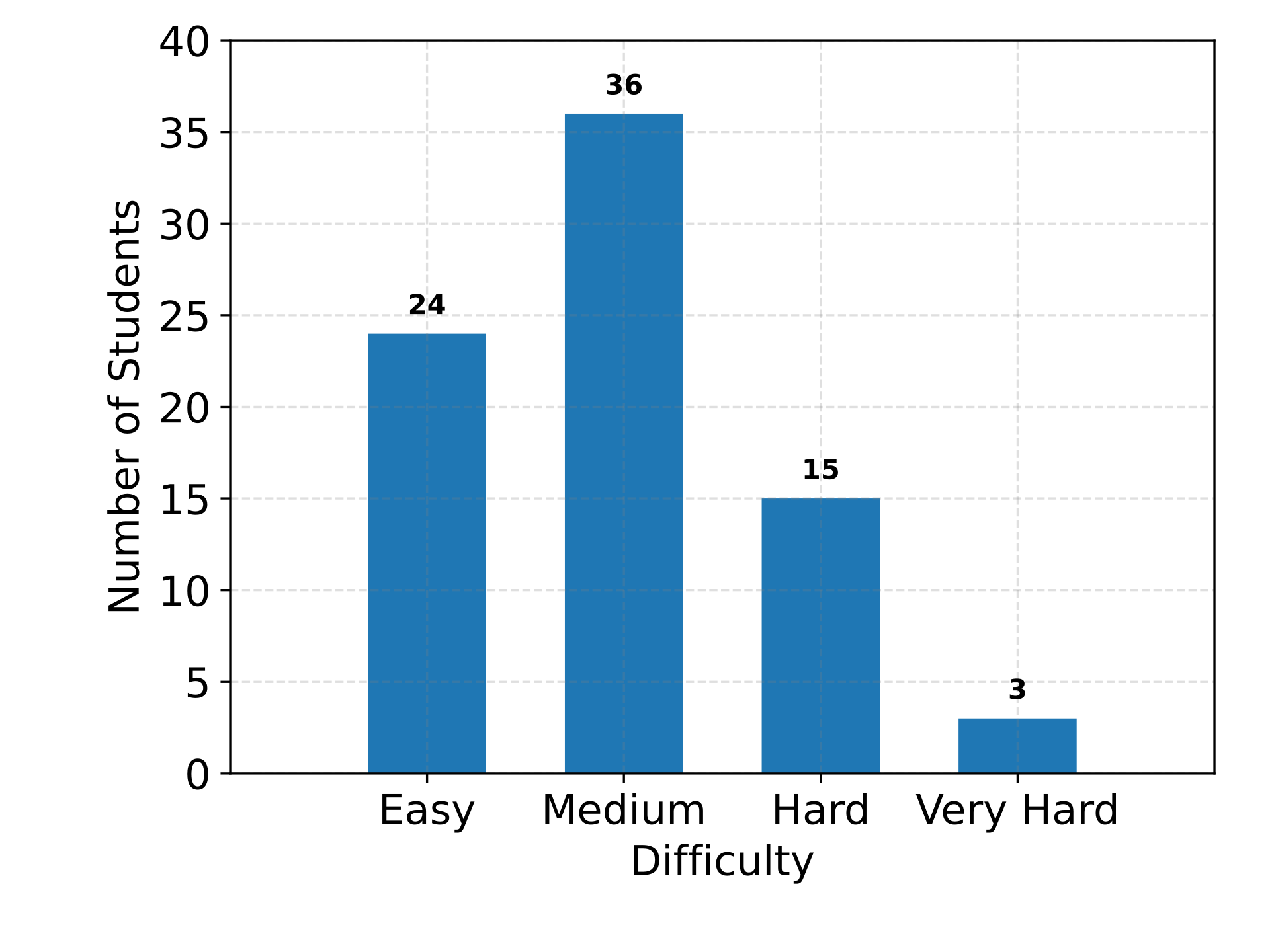}
		\caption{Students' perception of the difficulty level of the intermediate exam.}
		\label{intermediate-exam}
	\end{figure}

\begin{figure}[ht]
		\centering{\includegraphics[width=0.96\columnwidth]{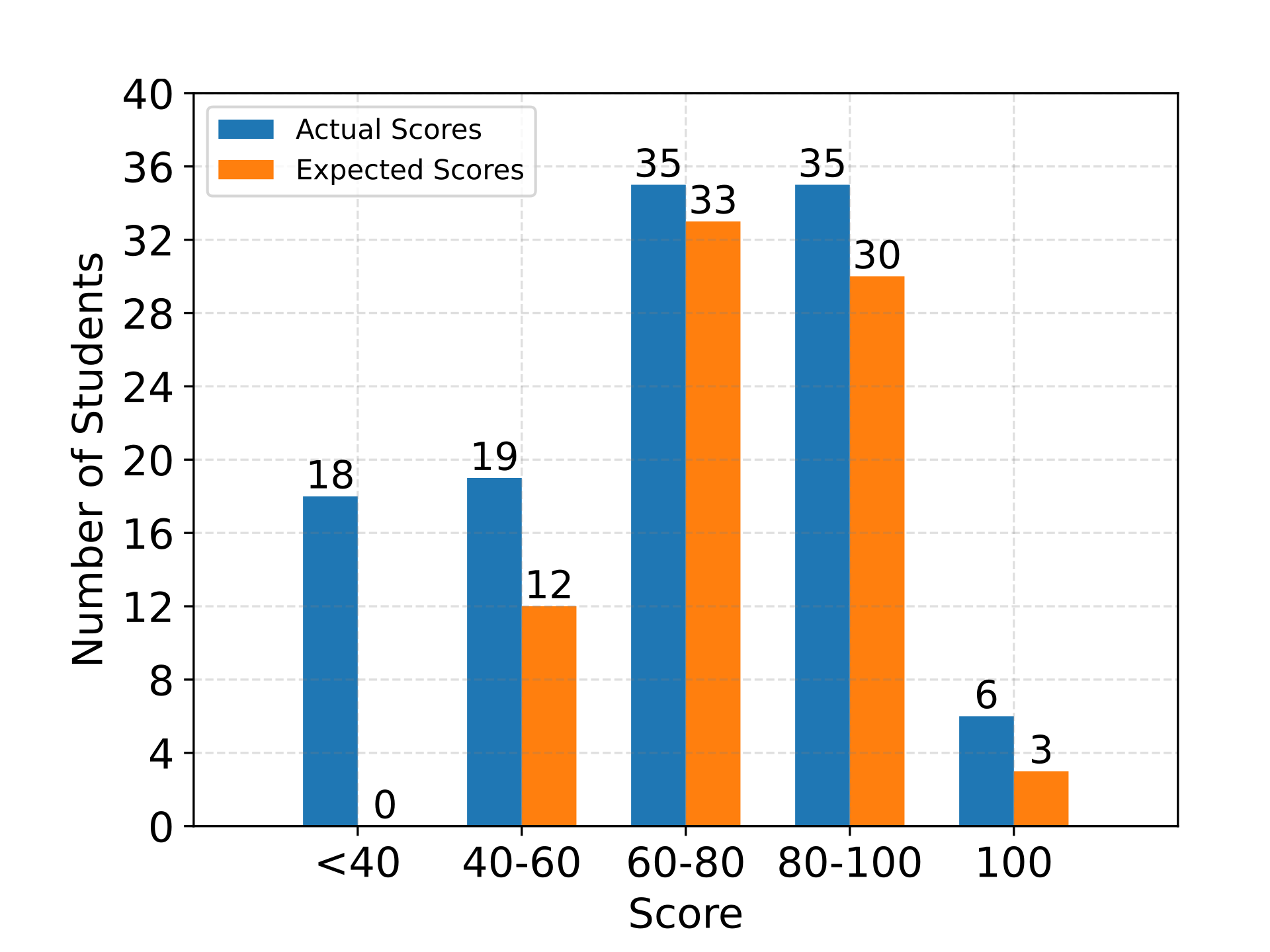}}
		\caption{Comparison between the expected score by the students (the percentage range that the student answered to the DiscordBot) with the real grade obtained by the students in the intermediate exam.}
		\label{intermediate-exam-score}
	\end{figure}

 Figure \ref{intermediate-exam} indicates that the intermediate exam presented a reasonable challenge for the students; however, more than half considered it to be a medium and easy difficulty level. The perceived difficulty of this exam was closely related to the students' performance. This is shown in Figure \ref{intermediate-exam-score}, where a pattern is evident between the expected and actual grades. This finding addresses RQ1: the difficulty level that students perceive in an activity (in this case, an evaluation) significantly correlates with the grades they achieve.
	
 Furthermore, the expectation of high student scores suggests that the teaching methodology is effective and that students are confident in their acquired knowledge throughout the course. This positive outcome reinforces the alignment between instructional strategies and student understanding.
		
	Figure \ref{intermediate-exam-score} shows grades lower than 40 \%, which was anticipated. Only about 9 \% of the students rated the exam as very difficult (Figure \ref{intermediate-exam}), and it is likely that these students did not respond to the question about their expected results (Figure \ref{intermediate-exam-score}), as it is natural to avoid predicting a poor grade.
	
\begin{figure}[ht]
		\centering{\includegraphics[width=0.96\columnwidth]{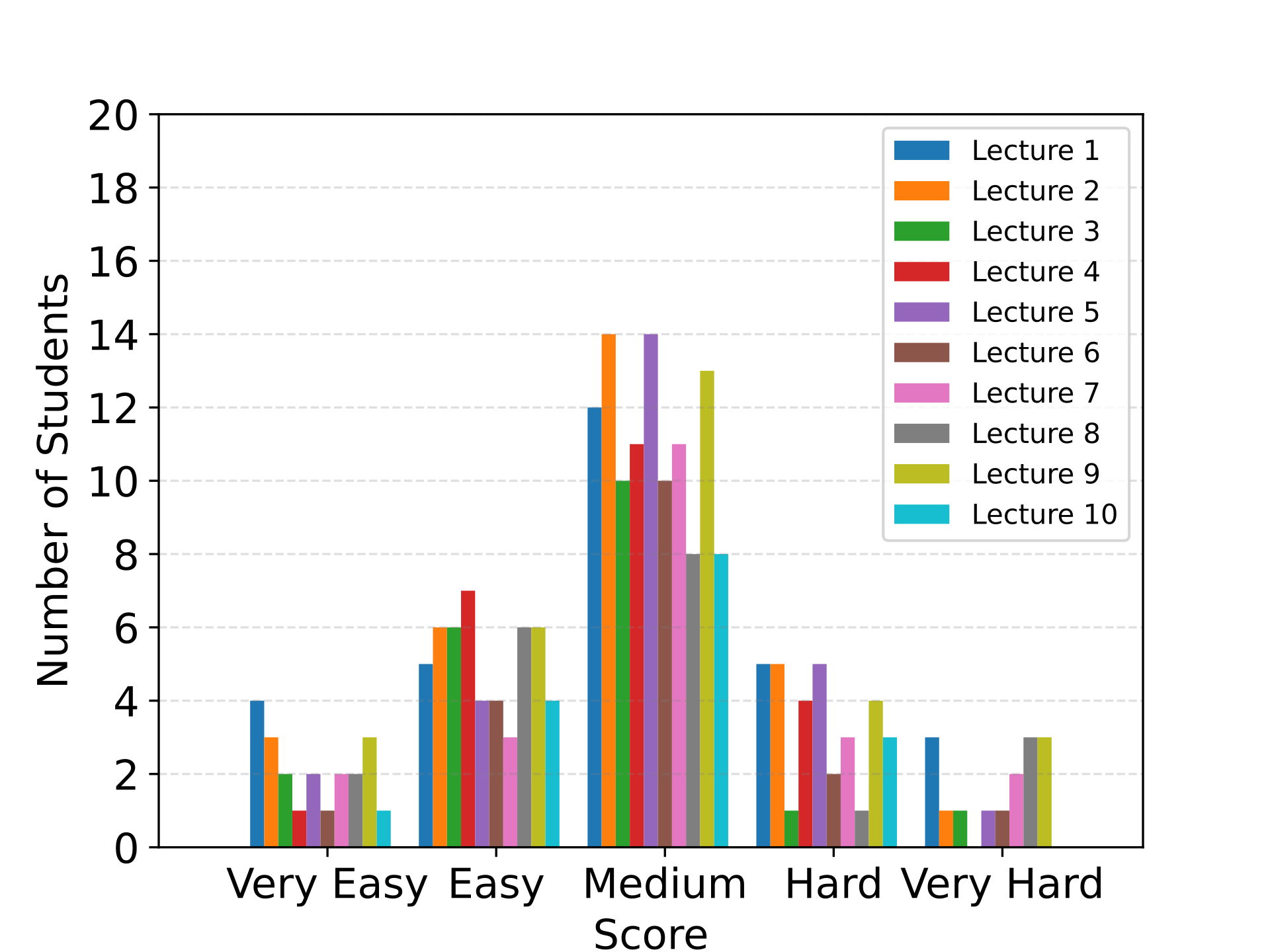}}
		\caption{Students' perception of the difficulty level of the lecture content.}
		\label{lecture-results}
	\end{figure}

	We also analyzed the students' perceptions about the lectures and quizzes, which led to important conclusions.
	As shown in Figure \ref{lecture-results}, in most cases, the lecture content has a medium or easy level of difficulty, which also corresponds to the intermediate exam results (Figure \ref{intermediate-exam-score}). 
	Nevertheless, many students find the lecture content hard or very hard, which we assume explains the grades lower than 40 \% (Figure \ref{intermediate-exam-score}).
	
\begin{figure}[ht]
		\centering{\includegraphics[width=0.96\columnwidth]{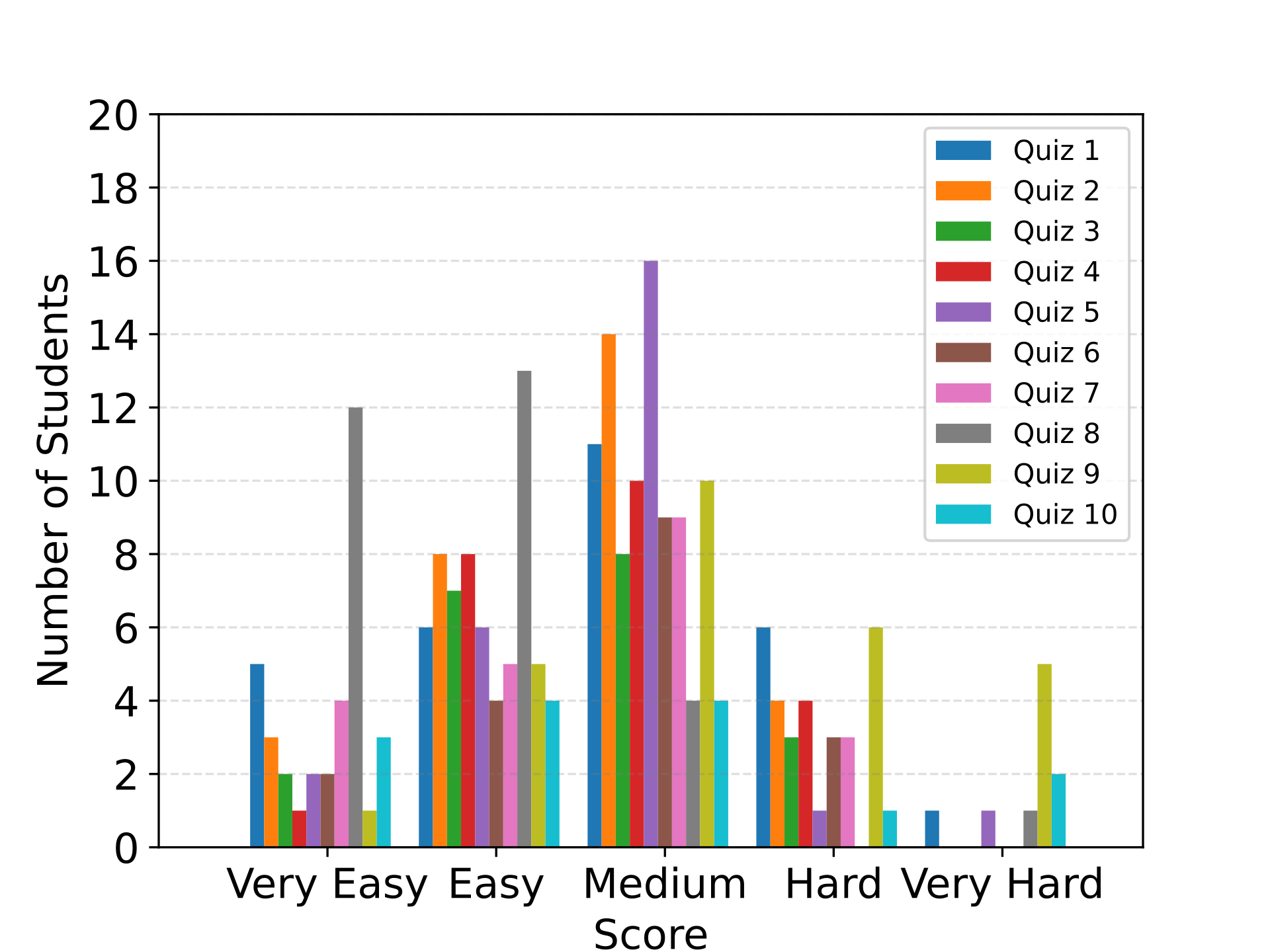}}
		\caption{Students' perception of the difficulty level of the quizzes.}
		\label{quiz-results}
	\end{figure}

	The Figures \ref{lecture-results} and \ref{quiz-results} indicated that the students' perceived difficulty level of the lectures' content is mirrored in the difficulty the students experienced when solving the corresponding quizzes (Figure \ref{quiz-results}). This finding addresses RQ2: the difficulty level perceived by the students about the lecture content is related to the difficulty level of the corresponding quiz. 
	
	Finally, the results obtained from the paper-base satisfaction survey about DiscordBot showed that 65.2 \% indicated that they regularly answered the surveys by the DiscordBot. Additionally, 69.6 \% also indicated that when they answered about their expected result of an activity, their actual grade was very close to it. According to the usability, 97.8 \% of the students indicated that the interaction was easy (50 \%) or normal (47.8 \%). Finally, 71.7 \% of students would like to use the DiscordBot in the coming semesters. These results showed us that DiscordBot is very highly accepted by students. 
	
	\section{Discussion}\label{D}
	
	The results of this study underscore the significant role that chatbots, specifically DiscordBot, can play in enhancing the experience in computing education with first-year students. The ability of DiscordBot to facilitate real-time feedback and interactive communication has increased student engagement and allowed instructors to tailor educational content dynamically in response to student needs. This adaptability is crucial in computing education, where technological advancement must be incorporated continuously. The findings indicate that DiscordBot has been particularly effective in managing course logistics and providing immediate support, which reduces administrative burdens and enhances the overall efficiency of the course management process.
	
	Despite these positive outcomes, several challenges remain. The reliance on manual data processing and the need for continuous monitoring of DiscordBot's performance highlights the importance of developing more autonomous and user-friendly systems. The collected feedback has also raised the question of how to effectively balance incorporating student inputs without compromising the structured progression of the course curriculum. Ensuring data privacy and preventing over-reliance are crucial ethical considerations when deploying and using chatbots.
	
	\section{Conclusion}\label{C}
	
	This study demonstrated that integrating a command-based chatbot into an introductory programming course on the Discord platform benefits students and instructors considerably. The DiscordBot helped to create a more responsive and engaging learning environment by facilitating continuous, real-time interactions and feedback. It has enabled instructors to swiftly adapt course content and tutorial methods to better suit the student's needs, thereby increasing the overall effectiveness of the educational process.
	
	Due to the utility of this DiscordBot, we intend to develop a graphical user interface (GUI) to eliminate the need for direct interaction with Python scripts for data processing. Integrating the GUI into the learning management system will prevent media breaks and simplify its use. This will enable instructors to modify bot configurations without manual edits. The main objective is to create a user-friendly chatbot that schools and universities can universally adopt to improve the learning process and student experience. This will help instructors to determine the acceptance level and incorporate new features based on student feedback.

	\bibliography{Paper_DiscordBot}

\begin{thebibliography}{10}

\bibitem{merelo2023chatbots}
J.~J. Merelo, P.~A. Castillo, A.~M. Mora, F.~Barranco, N.~Abbas, A.~Guill{\'e}n, and O.~Tsivitanidou, ``Chatbots and messaging platforms in the classroom: An analysis from the teacher’s perspective,'' {\em Education and Information Technologies}, pp.~1--36, 2023.

\bibitem{ramu2023generation}
D.~Ramu, R.~Jain, and A.~Jain, ``Generation z's ability to discriminate between ai-generated and human-authored text on discord,'' {\em arXiv:2401.04120}, 2023.

\bibitem{chempavathy2022ai}
B.~Chempavathy, S.~N. Prabhu, D.~Varshitha, Y.~Lokeswari, {\em et~al.}, ``Ai based chatbots using deep neural networks in education,'' in {\em 2nd International Conference on Artificial Intelligence and Smart Energy (ICAIS)}, pp.~124--130, IEEE, 2022.

\bibitem{adamopoulou2020overview}
E.~Adamopoulou and L.~Moussiades, ``An overview of chatbot technology,'' in {\em IFIP international conference on artificial intelligence applications and innovations}, pp.~373--383, Springer, 2020.

\bibitem{annuvs2023chatbots}
N.~Annu{\v{s}}, ``Chatbots in education: The impact of artificial intelligence based chatgpt on teachers and students,'' {\em International Journal of Advanced Natural Sciences and Engineering Researches}, vol.~7, no.~4, pp.~366--370, 2023.

\bibitem{riza2023use}
A.~N.~I. Riza, I.~Hidayah, and P.~I. Santosa, ``Use of chatbots in e-learning context: A systematic review,'' in {\em World AI IoT Congress}, pp.~0819--0824, IEEE, 2023.

\bibitem{adiguzel2023revolutionizing}
T.~Adiguzel, M.~H. Kaya, and F.~K. Cansu, ``Revolutionizing education with ai: Exploring the transformative potential of chatgpt,'' {\em Contemporary Educational Technology}, vol.~15, no.~3, p.~ep429, 2023.

\bibitem{kruglyk2020discord}
V.~Kruglyk, D.~Bukreiev, P.~Chornyi, E.~Kupchak, and A.~Sender, ``Discord platform as an online learning environment for emergencies,'' {\em Ukrainian Journal of Educational Studies and Information Technology}, vol.~8, no.~2, pp.~13--28, 2020.

\bibitem{lauricella2023examining}
S.~Lauricella, C.~Craig, and R.~Kay, ``Examining the benefits and challenges of using discord in online higher education classrooms,'' {\em Journal of Educational Informatics}, vol.~4, no.~2, pp.~20--31, 2023.

\bibitem{ayob2022promoting}
M.~A. Ayob, N.~A. Hadi, M.~E. H.~M. Pahroraji, B.~Ismail, and M.~N.~F. Saaid, ``{Promoting'Discord'as a Platform for Learning Engagement during COVID-19 Pandemic.},'' {\em Asian Journal of University Education}, vol.~18, no.~3, pp.~663--673, 2022.

\bibitem{vladoiu2020learning}
M.~Vladoiu and Z.~Constantinescu, ``{Learning during COVID-19 pandemic: Online education community, based on discord},'' in {\em 19th Conference on Networking in Education and Research (RoEduNet)}, pp.~1--6, IEEE, 2020.

\bibitem{jung2021developing}
H.~Jung and S.~Woo, ``Developing of facilitating-chatbot on computer supported collaborative learning and usability evaluation,'' {\em Journal of Digital Contents Society}, vol.~22, no.~7, pp.~1049--1057, 2021.

\bibitem{cp2021development}
G.~Mallikarjuna, A.~Srivastava, S.~Chakraborty, A.~Ghosh, and H.~Raj, ``Development of information technology telecom chatbot: An artificial intelligence and machine learning approach,'' in {\em 2nd International Conference on Intelligent Engineering and Management (ICIEM)}, pp.~216--221, IEEE, 2021.

\bibitem{krusche2017interactive}
S.~Krusche, A.~Seitz, J.~B{\"{o}}rstler, and B.~Br{\"{u}}gge, ``Interactive learning: Increasing student participation through shorter exercise cycles,'' in {\em 19th Australasian Computing Education Conference, {ACE}}, pp.~17--26, {ACM}.

\bibitem{krusche2020interactive}
S.~Krusche, N.~von Frankenberg, L.~M. Reimer, and B.~Bruegge, ``An interactive learning method to engage students in modeling,'' in {\em 42nd International Conference on Software Engineering (SEET)}, pp.~12--22, {ACM}, 2020.

\bibitem{krusche2023introduction}
S.~Krusche and J.~Berrezueta-Guzman, ``Introduction to programming using interactive learning,'' in {\em 35th International Conference on Software Engineering Education and Training (CSEE\&T)}, pp.~178--182, IEEE, 2023.

\bibitem{krusche2018artemis}
S.~Krusche and A.~Seitz, ``Artemis: An automatic assessment management system for interactive learning,'' in {\em 49th Technical Symposium on Computer Science Education, {SIGCSE}}, pp.~284--289, {ACM}, 2018.

\bibitem{linhuber2023constructive}
M.~Linhuber, J.~P. Bernius, and S.~Krusche, ``Constructive alignment in modern computing education: An open-source computer-based examination system,'' in {\em 23rd Koli Calling International Conference on Computing Education Research}, pp.~1--11, 2023.

\bibitem{fette2011websocket}
I.~Fette and A.~Melnikov, ``The websocket protocol,'' tech. rep., 2011.

\bibitem{wilson2022oauth}
Y.~Wilson and A.~Hingnikar, ``Oauth 2 and api authorization,'' in {\em Solving Identity Management in Modern Applications: Demystifying OAuth 2, OpenID Connect, and SAML 2}, pp.~63--101, Springer, 2022.

\end{thebibliography}
	\bibliographystyle{ieeetr}
	
\end{document}